\documentclass[10pt,twocolumn,letterpaper]{article}

\usepackage{iccv}
\usepackage{times}
\usepackage{epsfig}
\usepackage{graphicx}
\usepackage{amsmath}
\usepackage{amssymb}

\usepackage[accsupp]{axessibility}
\usepackage{array}
\newcolumntype{C}[1]{>{\centering\let\newline\\\arraybackslash\hspace{0pt}}m{#1}}

\usepackage{color, colortbl}
\definecolor{Gray}{gray}{0.9}
\usepackage{lipsum}
\usepackage{bbm}
\usepackage{verbatim}
\usepackage{multirow}

\usepackage{algpseudocode}
\usepackage{algorithm}
\usepackage{algorithmicx}

\usepackage{graphicx}

\usepackage{adjustbox}
\usepackage{comment}
\usepackage[normalem]{ulem}
\usepackage{bbm}
\usepackage{helvet}
\usepackage{tikz}
\usetikzlibrary{fadings,shapes.arrows,shadows}   
\tikzset{arrowstyle/.style={draw= black,single arrow,minimum height=#1, single arrow,
single arrow head extend=.4cm,align=center }}

\usepackage{amsthm}

\usepackage[breaklinks=true,bookmarks=false]{hyperref}
\iccvfinalcopy 

\def\iccvPaperID{6732} 
\def\httilde{\mbox{\tt\raisebox{-.5ex}{\symbol{126}}}}

\ificcvfinal\fi

\usepackage[capitalize]{cleveref}

\begin{document}

\title{Online Continual Learning on Hierarchical Label Expansion}

\author{Byung Hyun Lee$^{1,*}$, \ \ Okchul Jung$^{1,*}$, \ \ Jonghyun Choi$^{3,\dagger}$ and \ Se Young Chun$^{1,2,\dagger}$\\
$^1$Dept. of ECE, $^2$INMC \& IPAI, \ Seoul National University, \ Republic of Korea, \\
\ $^3$Yonsei University, \ Republic of Korea\\
{\tt\small \{ldlqudgus756,luckyjung96\}@snu.ac.kr} \ {\tt\small jc@yonsei.ac.kr} \ {\tt\small sychun@snu.ac.kr}}

\maketitle
\ificcvfinal\thispagestyle{empty}\fi

\begin{abstract}
Continual learning (CL) enables models to adapt to new tasks and environments without forgetting previously learned knowledge. While current CL setups have ignored the relationship between labels in the past task and the new task with or without small task overlaps, real-world scenarios often involve hierarchical relationships between old and new tasks, posing another challenge for traditional CL approaches. To address this challenge, we propose a novel multi-level hierarchical class incremental task configuration with an online learning constraint, called hierarchical label expansion (HLE). Our configuration allows a network to first learn coarse-grained classes, with data labels continually expanding to more fine-grained classes in various hierarchy depths. To tackle this new setup, we propose a rehearsal-based method that utilizes hierarchy-aware pseudo-labeling to incorporate hierarchical class information. Additionally, we propose a simple yet effective memory management and sampling strategy that selectively adopts samples of newly encountered classes. Our experiments demonstrate that our proposed method can effectively use hierarchy on our HLE setup to improve classification accuracy across all levels of hierarchies, regardless of depth and class imbalance ratio, outperforming prior state-of-the-art works by significant margins while also outperforming them on the conventional disjoint, blurry and i-Blurry CL setups.
\end{abstract}
\let\thefootnote\relax\footnotetext{$*$ Equal contribution, $\dagger$ Corresponding authors.}

\section{Introduction} 
\label{submission}
In real-world continual learning scenarios, new knowledge often augments existing understanding, typically following a hierarchical path from general to specific classes.
This hierarchical structure is not an anomaly, but rather an inherent part of many disciplines. The schema theory~\cite{brewer1981role, rumelhart1977representation} in cognitive psychology and the conceptual clustering theory~\cite{hofmann2001unsupervised} in machine learning both emphasize hierarchical organization of knowledge. The COBWEB algorithm~\cite{fisher1987knowledge}, a prominent machine learning method, uses hierarchical clustering for grouping related instances into meaningful categories.
Hierarchical organization is also observed in biology's taxonomy theory~\cite{bininda1999building}, classifying organisms based on shared traits, and in chemistry~\cite{heller2009iupac}, where elements are arranged hierarchically according to their atomic properties.
However, despite the prevalence of hierarchical relationships in these areas, many previous continual learning works~\cite{aljundi2019task, bang2021rainbow, bang2022online, koh2022online} do not fully incorporate these relationships. This may be an area that needs more attention, as hierarchical relationships could play a role in knowledge evolution in incremental learning.

\begin{figure*}[!t]
\begin{center}
\centerline{\includegraphics[width=\textwidth]{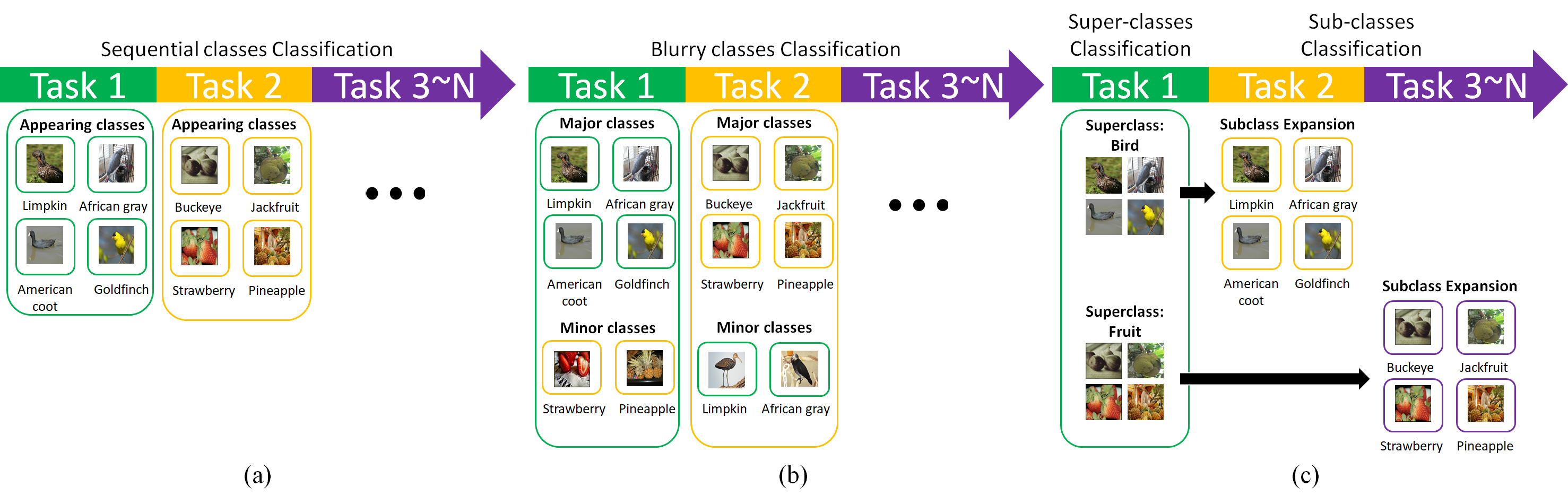}}
\caption{Comparison sketch between conventional, blurry, and our HLE setups. (a) Conventional task-free online CL setup gradually introduces new classes and classifies data without task identification (b) Blurry task-free online CL setup where classes are divided into major and minor categories at each task, with varying proportions, leads to unclear task boundaries (c) Proposed HLE CL setup features class label expansion where child class labels are added to parent class labels throughout the learning process.
}
\label{figure1:setup comparison}
\end{center}
\vskip -0.4in
\end{figure*}

Here we introduce a novel CL setup called Hierarchical Label Expansion (HLE), designed to account for hierarchical class relationships in task-free online CL. In HLE, class learning is incremental, with fine-grained classes derived from prior coarse-grained ones, effectively mirroring real-world knowledge accumulation.
As our proposed approach is designed for online continual learning, where data is seen only once in the data stream, each task's data is disjoint. We assess our models' performance using any-time inference~\cite{koh2022online} and evaluate classification accuracy for all levels of hierarchy. This demonstrates the potential of our approach to complement existing CL methods and enhance their evaluation. 
HLE encompasses both single and multiple hierarchy depths, as well as balanced and imbalanced class data scenarios. To tackle the CL on HLE, we propose a new CL method that utilizes pseudo-labeling based memory management (PL) and flexible memory sampling (FMS). 
This method effectively exploits hierarchy information between class labels in the dataset, resembling how knowledge is accumulated in real-world scenarios. Extensive experiments demonstrate that our approach outperforms state-of-the-art methods by substantial margins in HLE, while remaining superior in performance on existing CL setups including disjoint, blurry~\cite{bang2021rainbow} and i-Blurry~\cite{koh2022online}.

We summarize our contributions as follows:
\begin{enumerate}
\item We propose new online class-incremental, hierarchy-aware, task-free CL setups called HLE, designed to simulate how knowledge is accumulated in real-world scenarios.
\item We propose a new online CL method, {PL-FMS}, that consists of pseudo-labeling (PL) based memory management and flexible memory sampling (FMS) to better exploit hierarchy information and address the HLE setup.
\item We evaluate our approach on CIFAR100, Stanford-Cars, iNaturalist-19, and a novel dataset named ImageNet-Hier100, demonstrating that our method outperforms prior state-of-the-art works by significant margins on HLE while still outperforming them on the existing disjoint, blurry and i-Blurry CL setups.
\end{enumerate}

\section{Related Work}
\begin{figure*}[!t]
\begin{center}
\centerline{\includegraphics[width=1\textwidth]{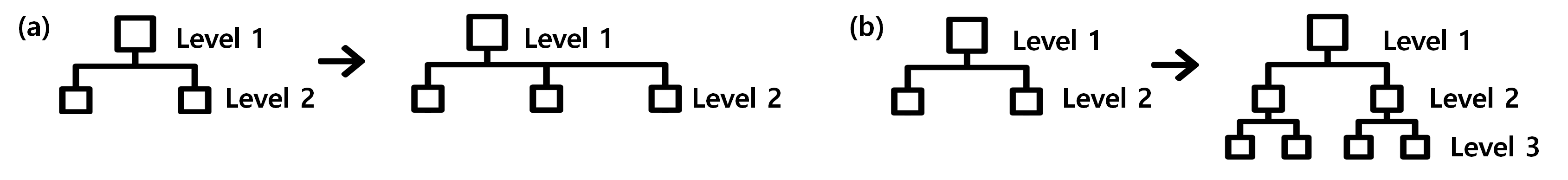}}
\caption{
An illustration of two HLE scenarios. (a) In single-depth scenario, fine-grained classes grow horizontally from coarse-grained ones within the same level. (b) In multiple-depth scenario, classes grow vertically from coarse to fine across different hierarchy levels.
}
\label{figure2:depth scenario}
\end{center}
\vskip -0.4in
\end{figure*}
\textbf{Continual learning setups.} 
Continual Learning (CL) setups can be classified into three categories: task-incremental, class-incremental, and domain-incremental learning setups~\cite{de2021continual, van2018three}.
Our work focuses on the class-incremental learning setting proposed by~\cite{rebuffi2017icarl}, where task identity is not given during inference, and the model is required to solve each task seen so far and infer which task it is presented with.
CL setups can be classified as either online~\cite{aljundi2019gradient, fini2020online, guo2022online, he2020incremental} or offline~\cite{ aljundi2018memory, chaudhry2018efficient, prabhu2020gdumb, rebuffi2017icarl, serra2018overcoming}. Our work focuses on the more challenging online CL setup where streamed samples are only used once, compared to the offline CL setup where data from each task can be used multiple times to train the model.
CL setups can also be categorized as task-free~\cite{al-shedivat2018continuous, aljundi2019task, li2019learn} or task-based~\cite{donti2017task, li2017learning, serra2018overcoming, shin2017continual}. 
Our work focuses on the former, where the model continuously learns and adapts to incoming data without explicit task information, unlike the latter where the model is informed about the tasks it must learn and adapt to.

Despite the considerable attention given to enhancing CL methods, their evaluation has been limited to rather restricted CL settings. To address this, novel CL setups with blurry task boundaries and corrupted labels in data stream~\cite{bang2021rainbow, bang2022online, koh2022online} have been proposed.     
A CL setup where classes are shared across tasks and presented sequentially as a stream with limited access to previous data was proposed by~\cite{bang2021rainbow}, while~\cite{bang2022online} suggested an online blurry CL setup with noisy labels. Recently, a new setup called `i-Blurry'~\cite{koh2022online} has been proposed, which combines the advantages of both blurry and disjoint setups by allowing continuous encounters of overlapping classes without suffering from restrictions of blurry and disjoint.
However, earlier works all assumed independent class labels, which is often not the case in reality. Our work proposes a complementary CL setup that models hierarchically correlating relationships between labels for online learning depicted in Figure~\ref{figure1:setup comparison}.

\textbf{Hierarchical classification.}
Various studies have utilized data's hierarchical structure to enhance tasks like image classification~\cite{bertinetto2020making, chang2021your, karthik2021no}, multi-label classification~\cite{wehrmann2018hierarchical}, object recognition~\cite{redmon2017yolo9000}, and semi-supervised approaches~\cite{garg2022hiermatch, wang2021hierarchical}. The hierarchical taxonomy is typically employed through label-embedding, hierarchical architecture-based, and hierarchical loss-based methods.

The label-embedding method maps class labels to vectors to represent semantic relationships and optimizes a loss on these embedded soft vectors.  
DeViSE~\cite{frome2013devise} maximizes the cosine similarity between image and label embeddings. It maps target classes to a unit hypersphere and penalizes the output that is more similar to false label embeddings using a ranking loss. 
Liu \textit{et al.}~\cite{liu2020hyperbolic} use hyperbolic geometry to learn hierarchical representations and minimize the Poincaré distance between Poincaré label embeddings and image feature embeddings, similar to DeViSE. 

Hierarchical architecture-based methods incorporate class hierarchy into the classifier architecture. 
Wu \textit{et al.}~\cite{wu2016learning} jointly optimize a multi-task loss function with cross-entropy loss applied at each hierarchy level. 
Redmon \textit{et al.}~\cite{redmon2017yolo9000} propose a probabilistic model, YOLOv2, for object detection and classification, with softmax applied at every coarse-category level to address the mutual exclusion of all classes in conventional softmax classifier. 
Chang \textit{et al.}~\cite{chang2021your} propose a multi-granularity classification architecture that uses level-specific classifiers to optimize fine-grained and coarse-grained recognition separately and improve fine-grained classification performance.

Hierarchical loss-based method incorporates hierarchical class relationships into the loss function and penalizes incorrect predictions while encouraging those that follow the hierarchy.
Deng \textit{et al.}~\cite{deng2010does} directly minimized the expected WordNet LCA height using kNN- and SVM-based classifiers, while Zhao \textit{et al.}~\cite{zhao2011large} modified multi-class logistic regression and added an `overlapping-group lasso penalty' to encourage the use of similar features for closely related classes.
Bertinetto \textit{et al.}~\cite{bertinetto2020making} proposed the hierarchical cross-entropy approach, where the loss function is based on conditional probabilities given parent-class probabilities. 

\section{Hierarchical Label Expansion}
In this section, we introduce our proposed HLE setup and present its configurations. Section~\ref{section 3.1} details the setup formulation, where the model is provided with samples only for the classes belonging to a single hierarchy level for each task. Section~\ref{section 3.2} describes the construction of single and multiple hierarchy depth scenarios in HLE to observe knowledge expansion at different levels.
\begin{figure*}[!ht]
\begin{center}
\centerline{\includegraphics[width=\textwidth]{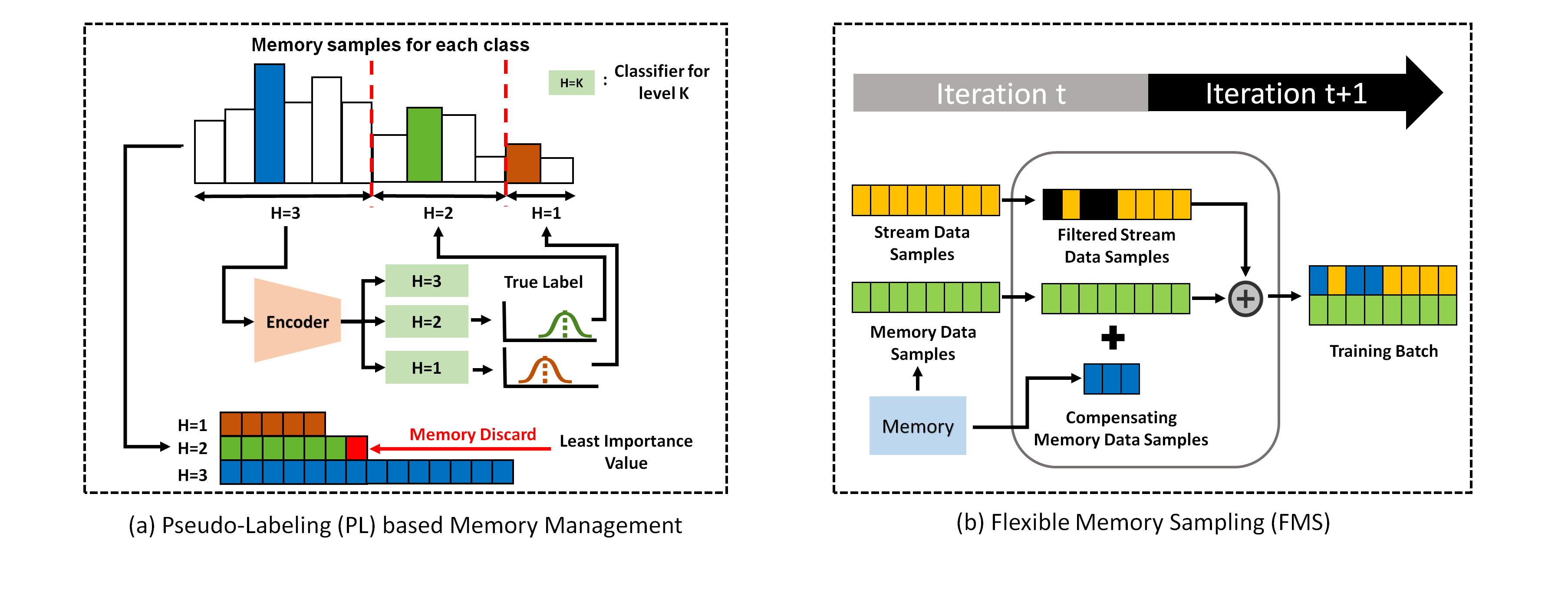}}
\vskip -0.2in
\caption{Sketch of our proposed method, PL-FMS's two components: PL and FMS. (a) Pseudo-Labeling based memory management (PL) outlines the method of discarding a data sample, which will be replaced with incoming data, based on its effect on reducing loss, irrespective of its label's nature (true or pseudo)..(b) Flexible Memory Sampling (FMS) shows formation of the training batch by filtering and compensating data samples.}
\label{figure3:PL-FMS}
\end{center}
\vskip -0.4in
\end{figure*}


\subsection{Hierarchical CL Configurations}\label{section 3.1}
Our HLE setup involves task-free online learning, where the model incrementally learns classes from various hierarchies both vertically and horizontally, agnostic to the task boundaries. The model is presumed to first learn coarse-grained classes, followed by fine-grained classes. Figure~\ref{figure1:setup comparison}(c) provides an overview of the HLE setup.

Formally, we consider the model encounters a stream of data points denoted by $\mathcal{T} = ((x_1, y_1), (x_2,y_2), \cdots)$, where $(x_j,y_j)$ is sampled from a data distribution $\mathcal{D}_{\mathbb{X} \times \mathbb{Y}}$, $x_j \in \mathbb{X}$ is the $j$th input (image) for the model, and $y_j \in \mathbb{Y}$ is the class label of $x_j$.
Often, the sequential tasks with the index $k$ can divide the data stream $\mathcal{T}$ into disjoint sub-sequences $\mathcal{T}_{1}, \mathcal{T}_{2}, \cdots$, where $\mathcal{T}_{k} = ((x_j, y_j) )_{j=t(k)}^{t(k+1)-1}$ and $t(k)$ is the start sample index for the $k$-th task. 
We define the class subset for the $k$-th task as $\mathbb{Y}_k \subseteq \mathbb{Y}$, which represents the set of classes that the model encounters during the $k$th-task.
The conventional CL usually assumes that the sampling distribution varies over time and the sampling distributions for tasks are mutually exclusive, \textit{i.e.}, $\mathbb{Y}_k \cap \mathbb{Y}_l = \emptyset$ for $k \neq l$.
However, there exist scenarios where more practical contexts need to be taken into account for reality. For example, the i-Blurry CL setup ~\cite{koh2022online} assumes that each task has both shared subset of classes  $\mathbb{Y}^{s}$, trained throughout the learning process, and disjoint subset of classes $\mathbb{Y}^{d}_{k}$, trained only at a specific task. For this case, the class subset $\mathbb{Y}_k$ is defined as $\mathbb{Y}_k = \mathbb{Y}^{s} \cup \mathbb{Y}^{d}_{k}$, which implies that $\mathbb{Y}_{k} \cap \mathbb{Y}_{l} = \mathbb{Y}^{s} \neq \emptyset$.

In a different direction to complement existing CL setups, our HLE allows more structures on $\mathbb{Y}$ by constructing a label relation between classes in $\mathbb{Y}$. Specifically, we consider that $\mathbb{Y}$ consists of classes from $H$ levels, so $\mathbb{Y} = \bigcup_{h=1}^{H} \mathbb{Y}^{h}$ and $\mathbb{Y}^{h} \cap \mathbb{Y}^{h^{'}} = \emptyset$ where $\mathbb{Y}^{h}$ is the label subset whose hierarchy level is $h$. 
By $h$, the smaller value of $h$ represents the hierarchy level for more coarse-grained classes.
In the HLE setup, each task conducts the label expansion for a subset of classes in level $h$ to their fine-grained classes in level $(h+1)$. That is, the labels are expanded by one level during a task. Let $\mathbb{Y}_{k}^{h} \subseteq \mathbb{Y}^{h}$ be the label subset for level $h$ that has been trained by the model until the $k$-th task. For the $(k+1)$-th task, a subset $\bar{\mathbb{Y}}_{k+1}^{h}$ of $\mathbb{Y}_{k}^{h}$ is selected to be newly expanded to a set of their fine-grained classes $\mathbb{Y}_{k,\text{new}}^{h+1}$, resulting in $\mathbb{Y}_{k} = \mathbb{Y}_{k,\text{new}}^{h+1}$.
To handle multiple hierarchy levels, our model consists of an encoder $f$ for feature embedding and multiple classifiers $\{g^{h}\}_{h=1}^{H}$, each corresponding to a hierarchy level. Specifically, $g^{h}(f(x))$ predicts the classes within level $h$ encountered until the current iteration. Regardless of its hierarchical position, each input is assigned a single label during training, and the model remains unaware of the hierarchy among classes. The hierarchy level is instead given as a soft hint to the model.
\subsection{Hierarchical CL Depth Scenarios}\label{section 3.2}
Our HLE setup includes two scenarios: single-depth and multiple-depth scenarios (existing setups are 0-depth), for hierarchical label expansion as depicted in Figure~\ref{figure2:depth scenario}. In the single-depth scenario, incremental learning is observed horizontally within the same hierarchy level, while in the multiple-depth scenario, new classes are introduced with increasing levels of specificity vertically. For the single-depth scenario, the model learns for all parent classes at the first task and partially expands them through subsequent tasks.
The single-depth scenario involves horizontal incremental learning within the same hierarchy level, starting with parent classes and broadening them in following tasks. This scenario is further explored through dual-label (overlapping data) and single-label (disjoint data) setups, as detailed in Table \ref{table1:label_scenario}.  
In the multiple-depth scenario, the model's ability to learn and expand hierarchical knowledge is tested while navigating complex vertical hierarchies by increasing the hierarchy level of classes to be learned for subsequent tasks, meaning that the model learns for classes of hierarchy level $h$ at the $h$th-task.

\section{Pseudo Labeling-based Flexible Memory Sampling (PL-FMS)}
In this section, we introduce our method which employs a rehearsal-based incremental learning approach, where models are trained using previously seen data from a stream buffer. Our method incorporates pseudo-labeling to fully utilize the hierarchical class relationship and a memory sampling strategy to flexibly build the training batch from stored and incoming data. Further details on our method's two main components, Pseudo-Labeling (PL) based Memory Management and Flexible Memory Sampling (FMS), are followed in sections \ref{PL} and \ref{FMS}, respectively.
\subsection{Pseudo-Labeling based Memory Management\label{PL}}


We introduce a novel memory management strategy that uses the model's predictions to generate pseudo-labels for each hierarchy level in our HLE setup, as shown in Figure \ref{figure3:PL-FMS}(a). 
This strategy is referred to as Pseudo-Labeling (PL) based memory management.

Basically, it first finds the modal label that are the most frequent in memory for class balance~\cite{bang2021rainbow, koh2022online, prabhu2020gdumb}.
Let $\mathcal{M}$ be the memory that stores samples from the data stream and $\mathcal{M}_{y} = \{(x_n, y_n) \in \mathcal{M} | y_n=y\}$ be the subset of the memory whose samples belong to the class $y$. 
For rehearsal-based method, we need to remove a sample from the memory to accept a new sample once $|\mathcal{M}|$ reaches the maximum memory size. 
To achieve this, we identify the class with the highest number of samples in the memory, which we denote as $\bar{y}=\text{arg} \max_{y}|\mathcal{M}_{y}|$. Prior works \cite{bang2021rainbow, koh2022online, prabhu2020gdumb} have typically removed samples only from $\mathcal{M}_{\bar{y}}$. To further improve the efficiency, we propose to consider samples from other classes hierarchically related to $\bar{y}$. To do so, we use the class probability predicted by the network, denoted as $p^{h}(x) = \sigma (g^{h}(f(x))) \in \mathbb{R}^{|\mathbb{Y}^{h}|}$ for level $h$, where $\sigma(\cdot)$ is the soft-max function.

We use the model to predict classes that are hierarchically related to $\bar{y}$. We do this by accumulating the model's predictions for samples in $\mathcal{M}_{\bar{y}}$ for all levels, except for the level of $\bar{y}$. The classes with the most predictions for each level are then identified, defined as:
\begin{equation}
\hat{y}^{h}(\mathcal{M}_{\bar{y}}) = \text{arg}\max_{y\in \mathbb{Y}^{h}} \sum_{(x, \bar{y}) \in \mathcal{M}_{\bar{y}} } \mathbf{1}_{y} (x),
\end{equation}
where $\mathbf{1}_{y}(x)$ is an indicator function defined as:
\begin{equation*}
    \mathbf{1}_{y}(x) =
        \begin{cases}
            \,\,\, 1, \,\,\, y = \text{arg}\max_{i} \,\, p_{i}^{h}(x) \\
            \,\,\, 0, \,\,\, \text{otherwise}.
        \end{cases}
\end{equation*}

In other words, the class at level $h$ that has the most predictions in $\mathcal{M}_{\bar{y}}$ is deemed as the class hierarchically related to $\bar{y}$.
By using the predicted classes for the other levels, we construct an index set of candidate samples to be removed from the memory as:

\begin{equation}
\mathcal{I}_{\bar{y}} = \{j | (x_j, y_j) \in \mathcal{M}_{\bar{y}} \cup \bigcup_{k=0, k\neq h}^{H} \mathcal{M}_{\hat{y}^{k}}  \}.
\end{equation}
To determine the index of a sample to remove, we adopt the sample-wise loss importance value, $\mathcal{H}_n$, introduced by ~\cite{koh2022online}. Specifically, $\mathcal{H}_n$ is computed as:
\begin{align*}
\mathcal{H}_n = L(\theta)-L(\theta_n),
\end{align*}
where $L(\theta)=\sum_{(x,y)\in\mathcal{M}} l(x,y;\theta)$ is the averaged loss in the memory and $\theta_n=\theta-\nabla_{\theta}l(x_n, y_n;\theta)$.
By using the loss importance value, we find the index $\hat{j}$ of the sample to remove whose measured importance is the least:
\begin{equation}
 \hat{j} = \text{arg} \min_{j\in \mathcal{I}_{\bar{y}}} \, \mathcal{H}_{j}.
\end{equation}
That is, we measure the decrease in loss for each sample during training and subsequently removes the data from the memory whose loss decrease is the least.

\begin{table*}[t!]
\centering
\scriptsize
\setlength{\tabcolsep}{1.7pt}
\renewcommand{\arraystretch}{1.4}
\begin{tabular}{ C{1.45cm} | C{1.2cm} C{1.2cm}  C{1.2cm} C{1.2cm}  C{1.2cm} C{1.2cm} | C{1.2cm} C{1.2cm}  C{1.2cm} C{1.2cm} C{1.2cm} C{1.2cm}}
\hline
\multirow{3}{*}{Methods}
                    & \multicolumn{6}{c|}{Single-Label Scenario} & \multicolumn{6}{c}{Dual-Label Scenario}  \\

                    & \multicolumn{2}{c}{CIFAR100} & \multicolumn{2}{c}{ImageNet-Hier100}  & \multicolumn{2}{c|}{Stanford Cars}  & \multicolumn{2}{c}{CIFAR100} & \multicolumn{2}{c}{ImageNet-Hier100}  & \multicolumn{2}{c}{Stanford Cars}  \\
                        
                   & H=1 & H=2 & H=1 & H=2 & H=1 & H=2 & H=1 & H=2 & H=1 & H=2 & H=1 & H=2  \\
 \hline

 ER  
  & 37.8$\pm$\scriptsize 2.06 & 31.3$\pm$\scriptsize 0.78
 & 73.4$\pm$\scriptsize 1.91 & 55.7$\pm$\scriptsize 1.87
 & 28.4$\pm$\scriptsize 0.73 & 4.01$\pm$\scriptsize 0.06
 & 42.0$\pm$\scriptsize 0.57 & 25.5$\pm$\scriptsize 0.33 
 & \underline{78.8$\pm$\scriptsize 0.82} & 57.2$\pm$\scriptsize 1.89
 & 37.8$\pm$\scriptsize 0.72 & 3.53$\pm$\scriptsize 0.44
 \\

 EWC++ 
 & 34.3$\pm$\scriptsize 0.68 & 27.1$\pm$\scriptsize 0.80
 & 73.4$\pm$\scriptsize 0.99 & 54.0$\pm$\scriptsize 1.34
 & 27.9$\pm$\scriptsize 0.74 & 3.42$\pm$\scriptsize 0.33
 & 39.9$\pm$\scriptsize 2.26 & 23.3$\pm$\scriptsize 1.93 
 & 76.3$\pm$\scriptsize 1.20 & 53.0$\pm$\scriptsize 3.32
 & 38.3$\pm$\scriptsize 0.47 & 3.17$\pm$\scriptsize 0.36
 \\

 BiC
 & 38.8$\pm$\scriptsize 0.41 & \underline{33.4$\pm$\scriptsize 1.41}
 & 72.5$\pm$\scriptsize 0.09 & 58.7$\pm$\scriptsize 0.78
 & 27.1$\pm$\scriptsize 1.08 & 3.05$\pm$\scriptsize 0.29
 & 42.1$\pm$\scriptsize 1.06 & 28.0$\pm$\scriptsize 1.01 
 & 77.7$\pm$\scriptsize 1.24 & 60.4$\pm$\scriptsize 0.30
 & 36.5$\pm$\scriptsize 1.04 & 3.26$\pm$\scriptsize 0.34
 \\

 MIR
 & 35.0$\pm$\scriptsize 1.47 & 28.6$\pm$\scriptsize 0.18
 & \underline{74.5$\pm$\scriptsize 0.90} & 57.3$\pm$\scriptsize 1.93
 & \underline{28.6$\pm$\scriptsize 1.09} & 4.50$\pm$\scriptsize 0.44 
 & 42.4$\pm$\scriptsize 0.95 & 26.2$\pm$\scriptsize 1.79
 & 78.5$\pm$\scriptsize 0.57 & 56.0$\pm$\scriptsize 2.25
 & \textbf{43.1$\pm$\scriptsize 1.18} & \underline{5.02$\pm$\scriptsize 0.74}
 \\

 RM
 & \underline{39.3$\pm$\scriptsize 0.83} & 25.9$\pm$\scriptsize 0.89
 & 69.7$\pm$\scriptsize 0.27 & \underline{61.0$\pm$\scriptsize 0.86}
 & 16.5$\pm$\scriptsize 4.05 & 2.83$\pm$\scriptsize 0.64
 & 38.2$\pm$\scriptsize 0.76 & 25.7$\pm$\scriptsize 1.12 
 & 71.5$\pm$\scriptsize 0.73 & \underline{63.1$\pm$\scriptsize 0.89}
 & 18.1$\pm$\scriptsize 2.54 & 3.29$\pm$\scriptsize 0.28
\\

 GDumb  
 & 26.2$\pm$\scriptsize 0.87 & 18.6$\pm$\scriptsize 0.09
 & 53.4$\pm$\scriptsize 1.18 & 37.2$\pm$\scriptsize 0.33
 & 16.6$\pm$\scriptsize 2.31 & 4.50$\pm$\scriptsize 0.12 
 & 25.7$\pm$\scriptsize 0.83 & 18.5$\pm$\scriptsize 1.11 
 & 59.2$\pm$\scriptsize 0.54 & 42.3$\pm$\scriptsize 0.54
 & 15.0$\pm$\scriptsize 1.40 & 4.06$\pm$\scriptsize 0.33
 \\
 
 CLIB  
 & 38.4$\pm$\scriptsize 0.58 & 32.6$\pm$\scriptsize 0.59
 & 64.6$\pm$\scriptsize 0.72 & 49.4$\pm$\scriptsize 1.32
 & 20.8$\pm$\scriptsize 2.08 & \underline{4.52$\pm$\scriptsize 0.78}
 & \underline{44.5$\pm$\scriptsize 0.87} & \underline{37.1$\pm$\scriptsize 0.20} 
 & 71.3$\pm$\scriptsize 0.76 & 55.4$\pm$\scriptsize 0.35
 & 19.1$\pm$\scriptsize 4.30 & 3.83$\pm$\scriptsize 0.78
 \\
 \hline

 PL-FMS    
 & \textbf{43.7$\pm$\scriptsize 0.13} & \textbf{36.4$\pm$\scriptsize 0.62}
 & \textbf{77.8$\pm$\scriptsize 1.32} & \textbf{64.6$\pm$\scriptsize 0.97}
 & \textbf{30.7$\pm$\scriptsize 4.39} & \textbf{13.2$\pm$\scriptsize 0.29}
 & \textbf{49.0$\pm$\scriptsize 0.19} & \textbf{39.5$\pm$\scriptsize 0.64}
 & \textbf{79.5$\pm$\scriptsize 0.54} & \textbf{67.2$\pm$\scriptsize 0.41}
 & \underline{42.0$\pm$\scriptsize 3.59} & \textbf{26.8$\pm$\scriptsize 3.27}
 \\
 \hline
\end{tabular}
\caption{
Experimental results of baseline methods and our proposed method evaluated on HLE setup for single-depth hierarchy scenario in CIFAR100, ImageNet-Hier100, and Stanford Cars. Dual-label means overlapping data between tasks, and single-label means disjoint data between tasks. Classification accuracy on hierarchy level 1 and 2 at the final task (\%) was measured for all datasets, and the results were averaged over three different random seeds.
}
\label{table1:label_scenario}
\end{table*}

\subsection{Flexible Memory Sampling (FMS) \label{FMS}}
Prior rehearsal-based methods~\cite{chrysakis2020online, he2020incremental, mir2019a, wu2019large} proposed directly including the stream buffer in training, leading to bias toward the data stream distribution and negatively impacting the model's performance. Using only memory samples for training was also suggested by~\cite{koh2022online}, but it limited adaptability to new classes. To balance the usage of memory and data stream, we propose Flexible Memory Sampling (FMS), a simple yet effective sampling strategy that flexibly adjusts the number of stream samples in the training batch. The approach is depicted in Figure \ref{figure3:PL-FMS} (b).

To construct a training batch $B_t$ at iteration $t$, ER utilizes all samples in the stream buffer $S_t$ and takes samples from the memory in an amount equal to $|S_t|$, which results in $|B_t| = 2 |S_t|$. 
Unlike ER, FMS randomly excludes samples from $S_t$ in the training process.
Let $T_c$ be the iteration when the class $c$ has been encountered for the first time. Then, we selectively include stream samples of class $c$ with increasing probability as $t- T_c$ gets larger, gradually adopting new classes from the stream buffer. In proportional to the value, the probability to include a stream sample of class $c$ is determined by a Bernoulli distribution for each class as:
\begin{equation}
\rho_{t}(c) \sim \text{Ber} \left( \text{min} \left( \frac{t-T_c}{T}, 1 \right) \right),
\label{eq_fms}
\end{equation}
where $T$ is a hyper-parameter that adjusts how fast the network adopts the stream samples for training.
Therefore, it resembles the memory-only training of ~\cite{koh2022online} immediately after encountering new classes, while it becomes more like the sampling approach of ER as $t-T_{c}$ gets larger. 
 
By combining those two strategies, we call our proposed method Pseudo Labeling-based Flexible Memory Sampling (PL-FMS). A detailed description of the algorithm for PL-FMS can be found in the supplementary material.

\section{Experiments}
\subsection{Experimental Setups}

\noindent\textbf{Datasets.}
We evaluate the Hierarchical Label Expansion (HLE) setup with a single-depth scenario on three datasets: \textbf{CIFAR100}~\cite{krizhevsky2009learning}, \textbf{Stanford Cars}~\cite{krause20133d}, and a newly constructed dataset called \textbf{ImageNet-Hier100}. 
CIFAR100 and Stanford Cars datasets each have 2 levels of hierarchy, with a total of (20,100) classes and (9,196) classes, respectively. The hierarchical taxonomy provided in each dataset was followed for the experiments. 
Additionally, we artificially constructed the ImageNet-Hier100, which is a subset of ImageNet~\cite{deng2009imagenet} based on the taxonomy of WordNet~\cite{miller1998wordnet}. This dataset also has 2 levels of hierarchy with a total of (10,100) classes.
Details on the curation of ImageNet data to construct ImageNet-Hier100 dataset are available in the supplementary material. 

We evaluate the HLE setup with a multiple-depth scenario on two datasets: \textbf{CIFAR100}~\cite{krizhevsky2009learning} and \textbf{iNaturalist-19}~\cite{van2018inaturalist}. 
For CIFAR100, we follow the hierarchical taxonomy as described in~\cite{garnot2020leveraging}, where the dataset has 5 levels of hierarchy with (2, 4, 8, 20, 100) classes, excluding the root node. 
For iNaturalist-19, we use the taxonomy in~\cite{bertinetto2020making}, where the dataset has 7 levels of hierarchy with (3, 4, 9, 34, 57, 72, 1010) classes, excluding the root node. 
Notably, only the iNaturalist-19 dataset is class-imbalanced among the two datasets.
Further details regarding the number of classes introduced at each task, dataset characteristics are available in the supplementary material. 

\noindent\textbf{Baselines.}
To provide a baseline for our method, we compare it with a range of previous works. 
We compare our rehearsal-based methods with previous works that were conducted under conventional CL setup, including \textbf{ER}~\cite{rolnick2019experience}, \textbf{EWC++}~\cite{chaudhry2018riemannian}, and \textbf{MIR}~\cite{mir2019a}. 
We also compare our rehearsal-based methods with works that were used in recently proposed CL setup, including \textbf{RM}~\cite{bang2021rainbow} and \textbf{CLIB}~\cite{koh2022online}. 
For regularization-based methods, we compare our methods with \textbf{BiC}~\cite{wu2019large} and \textbf{GDumb}~\cite{prabhu2020gdumb}. 
In the single-depth scenario, we evaluated all baseline methods, while in the multiple-depth scenario, we excluded MIR and GDumb as GDumb had the lowest performance and MIR had similar performance to ER, EWC++, and BiC. Further details about the experimental setup are available in the supplementary material.
\begin{figure}[t!]
\begin{center}
\centerline{\includegraphics[width=\columnwidth]{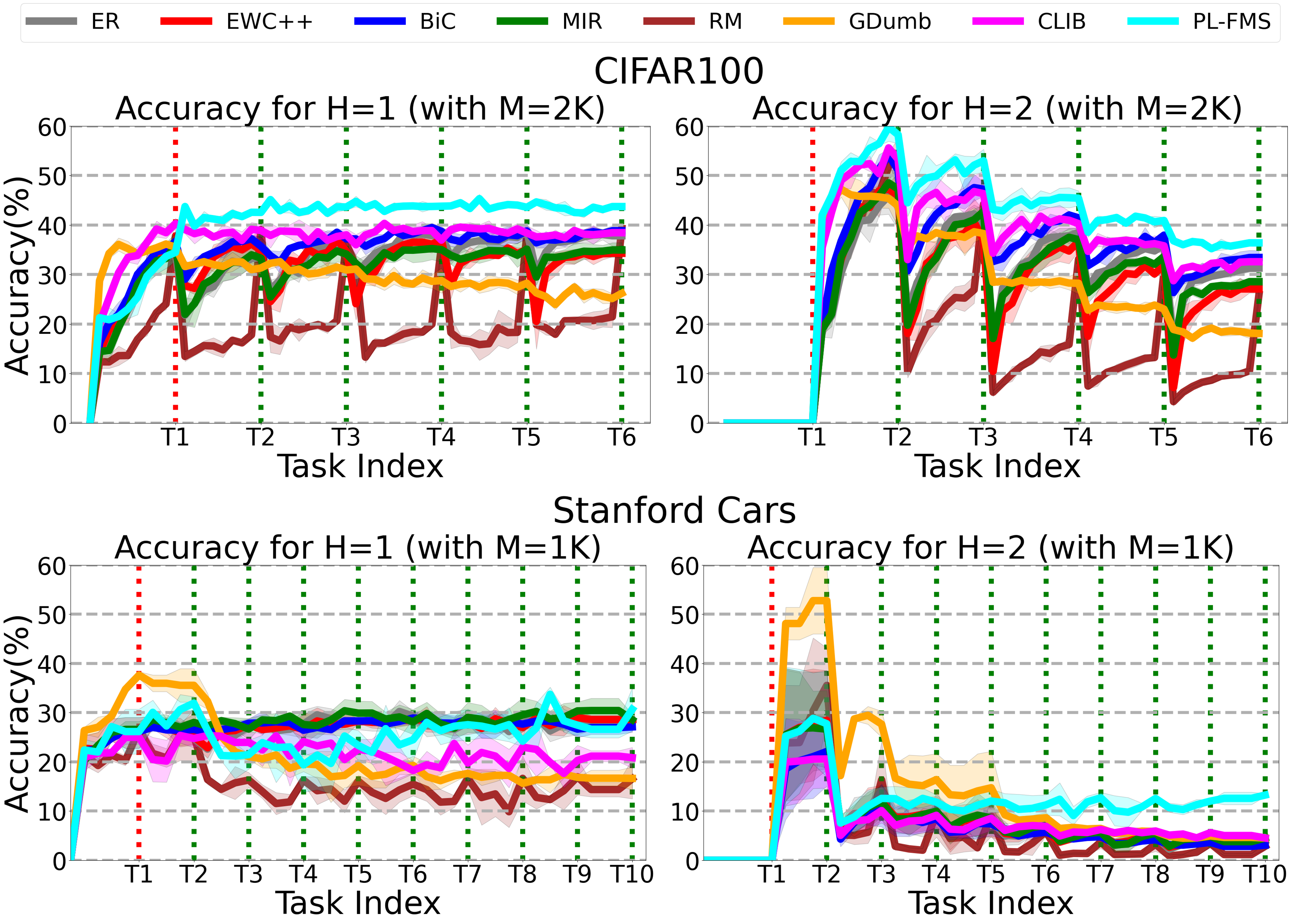}}
\caption{
Any-time inference results on CIFAR100 and Stanford Cars datasets for single-depth hierarchy. H=1 is parent classes and H=2 child classes. Task index 1 receives parent class labeled data and subsequent indexes receive child class labeled data. Each data point shows average accuracy over three runs ($\pm$ std. deviation).
}
\label{figure4:single_depth}
\end{center}
\vskip -0.4in
\end{figure}

\noindent\textbf{Scenarios.}
We conducted experiments in two scenarios: a single-depth hierarchy level and a multiple-depth hierarchy level, as detailed in Section~\ref{section 3.2} and illustrated in Figure~\ref{figure2:depth scenario}. Our HLE setup assumes disjoint data between tasks and is primarily evaluated under the single-label scenario. However, as described in Section~\ref{section 3.2}, we also conducted experiments under a dual-label scenario for the single-depth hierarchy level, where data had labels for both hierarchy levels.

\noindent\textbf{Evaluation metrics.}
We employ two primary evaluation metrics in our study: final classification accuracy for all hierarchy levels and any-time inference. 
Classification accuracy at the final task is a commonly used metric in evaluating continual learning methods, as demonstrated in previous works~\cite{chaudhry2018riemannian, han2018co, van2018three}. This metric measures the model's accuracy after all tasks have been learned as reported in the experimental tables. 
We also use any-time inference, as recommended in~\cite{koh2022online}, to assess the model's performance at any given time, crucial for observing knowledge expansion in our task-free setup. 
We report final accuracy in tables and any-time inference in figures for clarity over time. 
More details on these metrics are in the supplementary material.

\begin{table*}[t!]

\centering
\scriptsize
\setlength{\tabcolsep}{1.7pt}
\renewcommand{\arraystretch}{1.4}
\begin{tabular}{ C{1.45cm} | C{1.2cm} C{1.2cm}  C{1.2cm} C{1.2cm}  C{1.2cm} | C{1.2cm} C{1.2cm} C{1.2cm}  C{1.2cm} C{1.2cm} C{1.2cm} C{1.2cm}}
\hline
\multirow{2}{*}{Methods}
                    & \multicolumn{5}{c|}{CIFAR100} & \multicolumn{7}{c}{iNaturalist-19}  \\
 
            & H=1 & H=2 & H=3 & H=4 & H=5 & H=1 & H=2 & H=3 & H=4 & H=5 & H=6 & H=7 \\
\hline

 ER
 & 71.5$\pm$\scriptsize 4.44 & 58.4$\pm$\scriptsize 4.58 
 & 36.6$\pm$\scriptsize 4.78 & 18.1$\pm$\scriptsize 4.28
 & 7.47$\pm$\scriptsize 1.61 
 & 84.9$\pm$\scriptsize 6.03
 & \underline{84.9$\pm$\scriptsize 0.68} & 59.8$\pm$\scriptsize 15.5
 & 29.3$\pm$\scriptsize 3.28 & 17.8$\pm$\scriptsize 3.95
 & 13.0$\pm$\scriptsize 3.95 & 1.50$\pm$\scriptsize 0.77
 \\

 EWC++
 & 70.9$\pm$\scriptsize 2.83 & 56.6$\pm$\scriptsize 4.26
 & 35.8$\pm$\scriptsize 5.93 & 15.8$\pm$\scriptsize 3.94
 & 6.43$\pm$\scriptsize 1.28 
 & \textbf{87.4$\pm$\scriptsize 2.38}
 & 80.7$\pm$\scriptsize 1.19 & \underline{66.1$\pm$\scriptsize 9.80}
 & 29.4$\pm$\scriptsize 4.48 & 18.1$\pm$\scriptsize 6.53
 & 15.1$\pm$\scriptsize 5.73 & 1.88$\pm$\scriptsize 1.15
 \\

 BiC
 & 71.6$\pm$\scriptsize 1.01 & 63.5$\pm$\scriptsize 2.48
 & \underline{54.7$\pm$\scriptsize 0.61} & 33.8$\pm$\scriptsize 0.41
 & 19.8$\pm$\scriptsize 0.78
 & 79.5$\pm$\scriptsize 14.4
 & 76.3$\pm$\scriptsize 12.1 & 54.0$\pm$\scriptsize 27.4
 & 22.9$\pm$\scriptsize 10.3 & 14.8$\pm$\scriptsize 9.88
 & 11.2$\pm$\scriptsize 7.78 & 1.34$\pm$\scriptsize 1.41
 \\

 RM
 & \underline{74.2$\pm$\scriptsize 3.99} & \underline{65.0$\pm$\scriptsize 4.18}
 & 50.9$\pm$\scriptsize 1.40 & \underline{37.6$\pm$\scriptsize 0.60}
 & \underline{24.5$\pm$\scriptsize 2.54}
 & 74.0$\pm$\scriptsize 5.57
 & 69.7$\pm$\scriptsize 4.21 & 54.4$\pm$\scriptsize 2.20
 & 40.7$\pm$\scriptsize 1.15 & \underline{37.4$\pm$\scriptsize 0.85}
 & \underline{35.1$\pm$\scriptsize 0.44} & \underline{11.3$\pm$\scriptsize 0.33}
\\
 
 CLIB
 & 70.6$\pm$\scriptsize 4.05 & 59.5$\pm$\scriptsize 1.22
 & 47.6$\pm$\scriptsize 5.06 & 32.6$\pm$\scriptsize 1.76
 & 22.5$\pm$\scriptsize 2.08
 & \underline{87.2$\pm$\scriptsize 2.26}
 & 81.3$\pm$\scriptsize 4.78 & 62.4$\pm$\scriptsize 4.10
 & \underline{41.5$\pm$\scriptsize 0.97} & 35.3$\pm$\scriptsize 0.70
 & 33.2$\pm$\scriptsize 1.19 & 8.07$\pm$\scriptsize 0.94
 \\
 \hline
 
 PL-FMS    
 & \textbf{74.5$\pm$\scriptsize 4.63} & \textbf{65.6$\pm$\scriptsize 3.34} 
 & \textbf{56.0$\pm$\scriptsize 3.66} & \textbf{42.7$\pm$\scriptsize 1.79}
 & \textbf{30.8$\pm$\scriptsize 1.54}
 & 86.1$\pm$\scriptsize 3.15 
 & \textbf{88.4$\pm$\scriptsize 3.79}
 & \textbf{70.6$\pm$\scriptsize 3.17} 
 & \textbf{49.6$\pm$\scriptsize 2.42} 
 & \textbf{43.9$\pm$\scriptsize 1.86} 
 & \textbf{41.3$\pm$\scriptsize 2.57} 
 & \textbf{13.6$\pm$\scriptsize 0.28}
 \\
 
 \hline
\end{tabular}
\caption{
Experimental results reported for baseline methods and our proposed method evaluated on the HLE setup for the multiple-depth hierarchy scenario in CIFAR100 and iNaturalist-19. The classification accuracy on all hierarchy levels at the final task(\%) was measured for all datasets, and the results were averaged over three different random seeds.
}
\label{table2:multiple depth}
\end{table*}

\begin{figure*}[t!]
\begin{center}
\centerline{\includegraphics[width=\textwidth]{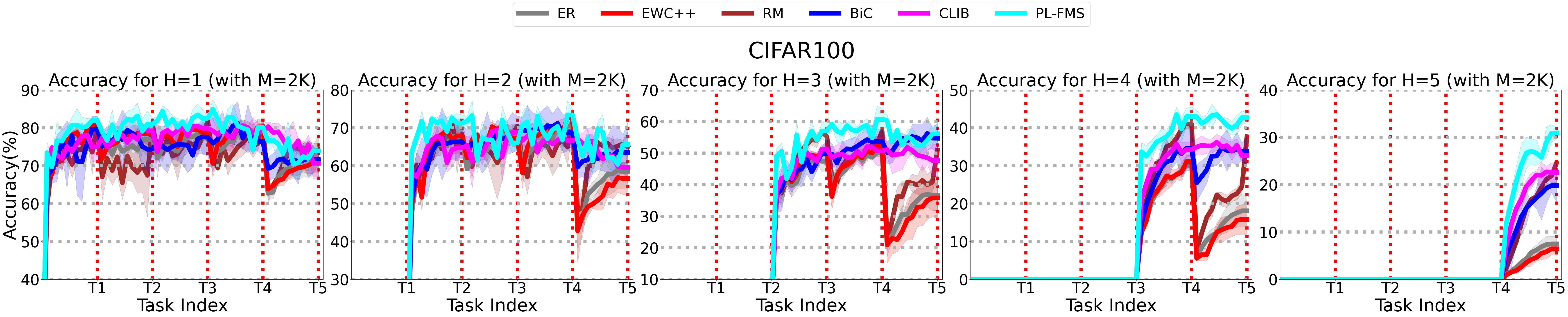}}
\caption{
Any-time inference results on CIFAR100 dataset for multiple-depth hierarchy. H=1 represents the coarsest level and H=5 represents the finest level of class hierarchy. The dotted line represents the point at which the model is fully given the task data for the corresponding task index. The reported data points represent the average accuracy over three runs ($\pm$ std. deviation).
}
\label{figure5:multiple depth}
\end{center}
\vskip -0.4in
\end{figure*}

\noindent\textbf{Implementation details.}
We implemented prior work using the~\cite{koh2022online} codebase, and applied AutoAugment~\cite{cubuk2019autoaugment} and CutMix~\cite{yun2019cutmix} as per their setup, but modified CutMix to mix samples only from the same hierarchy level to preserve the label distribution. 
We used ResNet34 as the base feature encoder across all methods, and adjusted batch sizes and update rates for each dataset: CIFAR100 (16, 3), ImageNet-Hier100 and iNaturalist-19 (64, 0.25), Stanford Cars (64, 0.5). 
Memory sizes were 1000, 2000, 5000, and 8000 for Stanford Cars, CIFAR100, ImageNet-Hier100, and iNaturalist-19, respectively. All methods except GDumb, CLIB, and PL-FMS used the Adam optimizer~\cite{kingma2014adam} with an initial learning rate of 0.0003 and an exponential learning rate scheduler. 
CLIB and our method used the same scheduler following the CLIB codebase. 
GDumb and CLIB adhered to their original optimization configurations.

\subsection{Single-Depth Scenario Analysis}
In the single-depth hierarchy scenario, knowledge expands horizontally within the same hierarchy level, as depicted in Figure~\ref{figure2:depth scenario}(a). The proposed HLE setup was evaluated on three datasets: CIFAR100 and ImageNet-Hier100, both class-balanced, and Stanford Cars, a class-imbalanced dataset, as reported in Table~\ref{table1:label_scenario} and Figure ~\ref{figure4:single_depth}.

Among baseline methods, GDumb showed the worst performance, while other methods showed varying performance depending on the dataset and hierarchy level. In CIFAR100, RM and BiC outperformed other baseline methods in hierarchy level 1 and 2, respectively. EWC++ and MIR demonstrated moderate performance in both hierarchy levels, while CLIB exhibited comparable performance to RM and BiC in hierarchy level 1. In ImageNet-Hier100, MIR showed the best performance in hierarchy level 1, while RM exhibited the best performance in hierarchy level 2. BiC showed moderate performance in hierarchy level 1, while EWC++ and ER demonstrated similar performance in hierarchy level 2. For Stanford Cars, MIR showed the best performance in hierarchy level 1, while CLIB performed well in hierarchy level 2. ER and BiC displayed similar performance in hierarchy level 1, while GDumb and RM exhibited the lowest and similar performance. In hierarchy level 2, all baseline methods showed similar performance, with overall accuracy between 3\% and 5\%. 
Our proposed method, PL-FMS, outperformed every baseline method in all single-label scenarios, with the largest improvement seen in the class-imbalanced dataset. It is worth noting that RM is a task-aware learning method that has demonstrated high performance under the HLE setup. This is achieved by a two-stage training approach, where the model is first trained on stream data samples and then fine-tuned using memory data samples resulting in an upsurge in performance near task boundaries. BiC includes a bias correction layer that effectively reduces dataset bias, but it does not directly improve performance near task boundaries. Additionally, MIR has shown significant performance by selecting high-loss importance samples, which helps to address the problem of catastrophic forgetting. 
However, GDumb consistently exhibits performance decay due to its fixed regularization coefficient, which limits its ability to adapt to new tasks.

\subsection{Multiple-Depth Scenario Analysis}
Our proposed HLE setup was evaluated on two datasets: class-balanced CIFAR100 and class-imbalanced iNaturalist-19, with the results reported in Table \ref{table2:multiple depth} and Figure~\ref{figure5:multiple depth}. The multiple-depth hierarchy scenario involves vertical knowledge expansion across all hierarchy levels, as shown in Figure~\ref{figure2:depth scenario} (b). All baseline methods were included except for GDumb and MIR. GDumb displayed consistently low performance across all datasets and hierarchy levels in single-depth hierarchy. MIR exhibited similar performance to that of ER and EWC++ in most cases, making it redundant to report separately.  

Our method, PL-FMS outperforms all baseline methods in CIFAR100, with the performance gap increasing significantly from hierarchy level 4 onwards, as reported in Table \ref{table2:multiple depth}. EWC++ had the lowest performance across all hierarchy levels, while ER performed similarly, but slightly better. RM and BiC had competing performances until hierarchy level 5. Throughout the hierarchy levels, CLIB's performance improved, ranking second among the baselines in the last hierarchy level. 
Note that most baseline methods suffer from catastrophic forgetting at all task indexes, but the most significant performance drop occurs at task boundary between task 4 and 5, as shown in Figure~\ref{figure5:multiple depth}.
This is due to the fact that the sampling strategy used by baseline methods for training batches fails to consider the biased class distribution induced by sub-categorization. On the other hand, PL-FMS and CLIB exhibit only a mild performance drop by avoiding direct adoption of the stream buffer.
PL-FMS outperformed all baseline methods in iNaturalist-19 except for level 1, with RM and CLIB showing the best performance in deeper hierarchy levels. EWC++ performed best only at the coarsest level and rapidly deteriorated thereafter, while BiC exhibited the worst performance overall. 
ER, EWC++, and BiC exhibited performance decline with increasing hierarchy levels, whereas RM and CLIB demonstrated significant performance improvements in comparison.

In Table~\ref{table2:multiple depth}, we observe a similar performance transition across the two datasets. However, at the hierarchy level 7, other baseline methods except for RM and CLIB show performance near 1\%, while RM, CLIB, and our method perform much better in the highest hierarchy level with performance above 10\%. We believe that ER, EWC++, and BiC exhibit significantly worse performance than RM, CLIB, and our method because they have not been tested under robust conditions, while RM and CLIB were proposed under more realistic conditions with blurry task boundaries and data streams. These methods are better equipped to deal with hierarchical knowledge formulation, which requires capturing common features throughout hierarchy trees. Overall, we observe that our method performs especially strongly under class imbalance situations, which is more similar to real-world scenarios. 

\begin{table}[h!]
\centering
\small
\setlength{\tabcolsep}{5pt}
\renewcommand{\arraystretch}{1.0}
\begin{tabular}{ C{1.75cm} | C{1.75cm} C{1.75cm}  C{1.75cm}}
\hline
                        
Methods            & Disjoint ~\cite{aljundi2019gradient} & Blurry ~\cite{bang2021rainbow} & i-Blurry ~\cite{koh2022online}  \\
 \hline

 ER
 & 36.6$\pm$\small 1.35
 & 24.5$\pm$\small 1.79 
 & 38.7$\pm$\small 0.51
 \\

 EWC++
 & 36.7$\pm$\small 1.04
 & 24.3$\pm$\small 1.20 
 & 38.7$\pm$\small 1.06 
 \\

 MIR
 & 34.5$\pm$\small 0.97
 & 24.0$\pm$\small 0.34 
 & 38.1$\pm$\small 0.69 
 \\

 RM
 & 35.4$\pm$\small 1.12
 & 37.8$\pm$\small 0.81
 & 36.7$\pm$\small 1.32
\\

 GDumb
 & 26.3$\pm$\small 0.43
 & 25.9$\pm$\small 0.08 
 & 32.1$\pm$\small 0.63
 \\

 CLIB
 & 38.0$\pm$\small 1.44
 & 38.3$\pm$\small 0.42
 & 43.4$\pm$\small 0.44 
 \\
 
 \hline
 FMS    
 & \textbf{39.2$\pm$\small 0.34}
 & \textbf{41.3$\pm$\small 1.98}
 & \textbf{45.3$\pm$\small 1.02}
 \\
 
 \hline
\end{tabular}
\caption{ 
Experimental results of baseline and FMS evaluated on three CL setups: conventional (disjoint), blurry, and i-Blurry. Test accuracy at the final task (\%) was measured for each setup and averaged over three runs with standard deviation reported.
}
\label{table3:3 prior setups}
\vskip -0.1in
\end{table}

\subsection{Label Regime Analysis}
Table~\ref{table1:label_scenario} presents the results of our experiment on a single-depth hierarchy, which we conducted under two scenarios: dual-label and single-label. 
Our dual-label scenario showed similar trends to the single-label scenario, with GDumb being the worst-performing method. Baseline methods that performed well in the single-label scenario had moderate performance in the dual-label scenario. Notably, incorporating the dual-label scenario resulted in an overall higher performance for the baseline methods in hierarchy level 1, although this was not consistent for hierarchy level 2 and varied among methods. Our proposed method, PL-FMS, consistently showed higher performance in the dual-label scenario across all datasets and hierarchy levels, suggesting that it is more adept at capturing hierarchy information in such scenarios, while still performing well in the single-label scenario against baseline methods.

\subsection{Prior CL Setups Analysis}
Table \ref{table3:3 prior setups} reports the results of our proposed HLE setup and baseline methods evaluated on various CL setups. Figure~\ref{figure1:setup comparison} depicts the difference between HLE and conventional CL setups. We evaluated the methods on disjoint, blurry~\cite{bang2021rainbow}, and i-Blurry~\cite{koh2022online} setups to check for code reproducibility and to observe whether our method could perform well on different setups. 
As reported in ~\cite{koh2022online}, CLIB exhibited superior or competitive performance to the other baseline methods across all previous setups, especially with large margin for the i-Blurry setup, since it has design for the i-Blurry setup. Note that our FMS outperformed CLIB for all the prior setups, which indicates that our method is not limited to the suggested HLE setup.

\begin{table}[h!]
\centering
\scriptsize
\setlength{\tabcolsep}{1.05pt}
\renewcommand{\arraystretch}{1.05}
\begin{tabular}{ C{1.5cm} | C{0.65cm} C{0.65cm} C{0.65cm} C{0.65cm} C{0.65cm} | C{0.65cm} C{0.65cm} | C{0.65cm} C{0.65cm} }

\hline
\multirow{2}{*}{Methods}
                    & \multicolumn{5}{c|}{Multiple-Depth} & \multicolumn{2}{c|}{Single-Label} & \multicolumn{2}{c}{Dual-Label}  \\

            & H=1 & H=2 & H=3 & H=4 & H=5 & H=1 & H=2 & H=1 & H=2 \\

\hline
 Proposed
 & \textbf{73.8}
 & \textbf{65.6}
 & \textbf{56.0}
 & \textbf{42.7}
 & \textbf{30.8}
 & \textbf{43.7}
 & \textbf{36.4}
 & \textbf{49.0}
 & \textbf{39.5}
 \\ 
 w/o PL
 & 73.5
 & 61.7
 & 48.2
 & 34.6 
 & 23.3
 & 41.3
 & 33.2 
 & 46.1
 & 33.9 
 \\
 w/o FMS
 & 71.4
 & 60.5
 & 45.9
 & 30.7
 & 21.5 
 & 39.6
 & 31.5
 & 43.8
 & 32.8
 \\
 \hline

\end{tabular}
\caption{
Ablation study conducted on CIFAR100 to compare the performance of PL-FMS, with and without the PL and FMS components, as well as their combination. Average accuracy across three runs is reported.
}
\label{table4:ablation study}
\vskip -0.05in
\end{table}

\subsection{Ablation Study}
In Table~\ref{table4:ablation study}, we conducted an ablation study to determine the contribution of each component in our proposed method for multi-depth, single-label, and dual label scenarios. The two components, PL and FMS, were evaluated separately to observe the performance gain achieved by each component. Results indicate that PL contributes more to the overall performance gain compared to FMS. However, when used together, the two components benefit each other and show higher performance gain for all scenarios.
\begin{table}[h!]
\centering
\scriptsize
\setlength{\tabcolsep}{0.75pt}
\renewcommand{\arraystretch}{1.05}
\begin{tabular}{ C{1.1cm} | C{1.35cm} C{1.35cm} C{1.35cm} C{1.35cm} C{1.35cm} }
\hline
\multirow{2}{*}{Methods}
                    & \multicolumn{5}{c}{Multiple-Depth} \\

            & H=1 & H=2 & H=3 & H=4 & H=5 \\
\hline
 Oracle
 & 94.6$\pm$\scriptsize 0.7 
 & 92.3$\pm$\scriptsize 1.2 
 & 84.1$\pm$\scriptsize 0.8 
 & 73.5$\pm$\scriptsize 1.2 
  & 61.1$\pm$\scriptsize 1.4 
 \\
PL-FMS-T
 & 87.7$\pm$\scriptsize 0.2 
 & 81.9$\pm$\scriptsize 0.5 
 & 69.0$\pm$\scriptsize 1.7 
 & 51.5$\pm$\scriptsize 2.0 
 & 35.2$\pm$\scriptsize 1.3 
 \\
PL-FMS
 & 74.5$\pm$\scriptsize 4.6
 & 65.6$\pm$\scriptsize 3.3
 & 56.0$\pm$\scriptsize 3.7
 & 42.7$\pm$\scriptsize 1.8
 & 30.8$\pm$\scriptsize 1.5
 \\
 \hline

\end{tabular}
\caption{CIFAR100 multiple-depth scenario results (\%) across three runs. `Oracle': All-classes-at-once (offline batch learning, assuming unlimited access to true class hierarchy labels during training). `PL-FMS-T': PL-FMS with true class hierarchy labels.}
\label{table 1:All class hierarchy label setting}
\vskip -0.05in
\end{table}

We also compared our method against an oracle result obtained via offline batch learning on all classes simultaneously and an approach leveraging true class hierarchy labels (PL-FMS-T). As seen in Table~\ref{table 1:All class hierarchy label setting}, our method gains from scenarios where true class hierarchy is available.

\begin{table}[h!]
\centering
\scriptsize
\setlength{\tabcolsep}{1.05pt}
\renewcommand{\arraystretch}{1.05}
\begin{tabular}{ C{1.5cm} | C{0.65cm} C{0.65cm} C{0.65cm} C{0.65cm} C{0.65cm} | C{0.65cm} C{0.65cm} | C{0.65cm} C{0.65cm} }

\hline
\multirow{2}{*}{Methods}
                    & \multicolumn{5}{c|}{Multiple-Depth} & \multicolumn{2}{c|}{Single-Label} & \multicolumn{2}{c}{Dual-Label}  \\

            & H=1 & H=2 & H=3 & H=4 & H=5 & H=1 & H=2 & H=1 & H=2 \\

\hline
 T=500
 & 79.0
 & 62.4
 & 52.1
 & 37.2
 & 27.4
 & 40.0 
 & 32.8
 & 47.6 
 & 36.7
 \\ 
 T=1,500
 & 73.0
 & 65.4
 & 50.2
 & 36.4
 & 28.7 
 & 38.3
 & 32.8
 & 46.6
 & 36.9
 \\
 T=5,000
 & \textbf{74.5}
 & \textbf{65.6} 
 & \textbf{56.0}
 & \textbf{42.7}
 & \textbf{30.8}
 & \textbf{43.7}
 & \textbf{36.4}
 & \textbf{49.0}
 & \textbf{39.5}
 \\
 T=15,000
 & 72.7
 & 64.7 
 & 50.5 
 & 34.6
 & 28.3
 & 41.4
 & 33.5
 & 48.8
 & 38.6
 \\
 T=50,000
 & 77.9 
 & 67.0
 & 52.9
 & 39.0
 & 28.8
 & 39.0
 & 31.7
 & 46.3
 & 34.4
 \\
 \hline

\end{tabular}
\caption{Effect of hyperparameter $T$ in Eq.~\ref{eq_fms} (\%) of PL-FMS on CIFAR100.}
\label{table 3:hyperparameterization study}
\vskip -0.05in
\end{table}

Table~\ref{table 3:hyperparameterization study} shows how the hyperparameter $T$ in PL-FMS, controlling the network's adaptation speed during training, affects performance. Choosing a value of 5,000 for $T$ yielded the highest accuracy, especially in fine-grained hierarchy classes across all scenarios.

\section{Conclusion}
In this work, we propose hierarchical label expansion (HLE), novel hierarchical class incremental task configurations with an online learning constraint, that complement existing CL setups by mimicking knowledge expansion. Then, we propose Pseudo-Labeling (PL) based memory management and Flexible Memory Sampling (FMS) to tackle this newly proposed CL setups for fully exploiting the inherent data hierarchy.  
Our proposed method outperforms prior state-of-the-art works by significant margins on our HLE setups across all levels of hierarchies, regardless of depth and class imbalance while also outperforming them on the previous disjoint, blurry and i-Blurry CL setups.

\section*{Acknowledgments}
\noindent
This work was partly supported by the National Research Foundation of Korea(NRF) grants funded by the Korea government (MSIT) (NRF-2022R1A4A1030579, NRF-2022M3C1A309202211, NRF-2022R1A2C4002300 5\%), IITP grants (No.2020-0-01361, AI GS Program (Yonsei University) 5\%, No.2021-0-02068, AI Innovation Hub 5\%, 2022-0-00077 5\%, 2022-0-00113 5\%, 2022-0-00959 5\%) funded by the Korea government (MSIT), and Creative-Pioneering Researchers Program through Seoul National University. Also, the authors acknowledged the financial support from the BK21 FOUR program of the Education and Research Program for Future ICT Pioneers, Seoul National University.

{\small
\bibliographystyle{ieee_fullname}
\bibliography{hierarchicalCL}
}

\clearpage
\appendix

\twocolumn[{%
\renewcommand\twocolumn[1][]{#1}%

\begin{center}
\bigskip 
\bigskip 
\textbf{\Large Online Continual Learning on Hierarchical Label Expansion \\ \emph{Supplementary Material} \\}
\bigskip 
\bigskip 
{\large Byung Hyun Lee$^{1,*}$, \ \ Okchul Jung$^{1,*}$, \ \ Jonghyun Choi$^{3,\dagger}$ and \ Se Young Chun$^{1,2,\dagger}$\\
$^1$Dept. of ECE, $^2$INMC \& IPAI, \ Seoul National University, \ Republic of Korea, \\
\ $^3$Yonsei University, \ Republic of Korea\\
{\tt\small \{ldlqudgus756,luckyjung96\}@snu.ac.kr} \ {\tt\small jc@yonsei.ac.kr} \ {\tt\small sychun@snu.ac.kr}
}
\bigskip 
\bigskip 

\maketitle

\setcounter{equation}{0} 
\setcounter{figure}{0} 
\setcounter{table}{0} 
\setcounter{page}{1} 
\makeatletter 

\renewcommand{\thefigure}{S.\arabic{figure}}
\renewcommand{\thetable}{S.\arabic{table}}
\renewcommand{\thesection}{S.\arabic{section}}

\end{center}%
}
]
\let\thefootnote\relax\footnotetext{$*$  Equal contribution, $\dagger$ Corresponding author.}

\crefname{section}{Sec.}{Secs.}
\Crefname{section}{Section}{Sections}
\Crefname{table}{Table}{Tables}
\crefname{table}{Tab.}{Tabs.}
\renewcommand{\thesection}{S\arabic{section}}
\renewcommand{\thefigure}{S\arabic{figure}}
\renewcommand{\thetable}{S\arabic{figure}}
\renewcommand{\theequation}{S\arabic{figure}}

\iccvfinalcopy 

\def\iccvPaperID{6732} 
\def\httilde{\mbox{\tt\raisebox{-.5ex}{\symbol{126}}}}

\section{Algorithms of Pseudo-Labeling based Flexible Memory Sampling (PL-FMS)}
In this section, we offer a detailed explanation of the Pseudo-Labeling based Flexible Memory Sampling. The details for pseudo-labeling based memory management and flexible memory sampling can be found in Alg. \ref{plmm} and Alg. \ref{fms} respectively. It is important to note that both algorithms are executed for every iteration, ensuring that the sampling process is constantly optimized. In Alg \ref{fms}, when $\rho_{t}(y)$ is not equal to $1$, then the sample from the stream data samples $\mathcal{S}_t$ is rejected and a sample is randomly selected from the set difference of the memory and the already selected memory samples. Overall, our Pseudo-Labeling based Flexible Memory Sampling approach is designed to improve the efficiency and effectiveness of the sampling process. 

\begin{algorithm}
\caption{Pseudo-Labeling based Memory Management}
\label{plmm}
\resizebox{\linewidth}{!}{
\begin{minipage}{\linewidth}
\begin{algorithmic}
\State \textbf{Input} levels of hierarchy $H$, feature extractor $\mathcal{F}$, classifiers $\{\mathcal{G}^{k}\}_{k=1}^{H}$, softmax function $\sigma$, memory $\mathcal{M}$, memory size $m$, label sets for hierarchy levels $\{\mathbb{Y}^{k}\}_{k=1}^{H}$, new sample $(x_{new}, y_{new})$, hierarchy level of the new sample $h$, per-sample importance measures for samples $\mathcal{H}$
\If{$|\mathcal{M}| < m$}
    \State $\mathcal{M} \gets \mathcal{M} \cup \{(x_{new}, y_{new})\}$
\Else
    \State $\bar{y} = \text{arg} \max_{y} | \{ (x_n,y_n) \in \mathcal{M} | y_n=y \}| $
    \State $\mathcal{M}_{\bar{y}} = \{(x_n, y_n) \in \mathcal{M} | y_n=\bar{y}\} $
    \State $\mathcal{I}_{\bar{y}} = \{j | (x_j, y_j) \in \mathcal{M}_{\bar{y}} \}$
    \For{$k=1,2,\cdots,H$}
        \If{$k \neq h$}
            \State $s = (0,0,\cdots,0) \in \mathbb{R}^{|\mathbb{Y}^{k}|}$
            
            \For{ $(x,y) \in \mathcal{M}_{\bar{y}}$ }
                \State $\hat{i} = \text{arg}\max_{i} \,\, \sigma (\mathcal{G}^{k} (\mathcal{F}(x)))_{i}$
                \State $s_{\hat{i}} \gets s_{\hat{i}}+1$
            \EndFor
            \State $\hat{y}^{k} = \text{arg} \max_{i\in\mathbb{Y}^{k}} \,\, s_{i} $
            \State $\mathcal{I}_{\bar{y}} \gets \mathcal{I}_{\bar{y}} \cup \{j|(x_j, y_j) \in \mathcal{M} , y_j=\hat{y}^{k}\}$ 
        \EndIf
    \EndFor
    \State $\hat{j} = \text{arg}\min_{j\in\mathcal{I}_{\bar{y}}} \mathcal{H}_{j} $
    \State $\mathcal{M} \gets (\mathcal{M} \backslash \{(x_{\hat{j}}, y_{\hat{j}})\}) \cup \{ (x_{new}, y_{new}) \}$
\EndIf
\State \textbf{Output} $\mathcal{M}$
\end{algorithmic}
\end{minipage}
}
\end{algorithm}

\begin{algorithm}
\caption{Flexible Memory Sampling}
\label{fms}
\resizebox{\linewidth}{!}{
\begin{minipage}{\linewidth}
\begin{algorithmic}
\State \textbf{Input} memory $\mathcal{M}$, training iteration $t$, iterations encountering class $y$ at the first time $T_{y}$, normalizing factor $T$, stream data samples $\mathcal{S}_t$, memory data samples $\mathcal{N}_t$
\State $\mathcal{B}_t = \mathcal{N}_t$
\For{$(x,y) \in \mathcal{S}_t$}
    \State $\rho_{t}(y) \sim \text{Bern} \left( \text{min} \left( \frac{t-T_y}{T}, 1 \right) \right)$
    \If{$\rho_{t}(y)$ is not $1$}
        \State $(x', y') \sim \mathcal{U}_{\mathcal{M}\backslash \mathcal{B}_t}$  
        \State $\backslash \backslash \,\, \mathcal{U}_A$ : uniform random sampler over a set $A$
        \State $\mathcal{B}_t \gets \mathcal{B}_t \cup \{ (x',y') \}$
    \Else
        \State $\mathcal{B}_t \gets \mathcal{B}_t \cup \{ (x,y) \}$
    \EndIf
\EndFor
\State \textbf{Output} $\mathcal{B}_t$

\end{algorithmic}
\end{minipage}
}
\end{algorithm}

\section{Details on Dataset Configuration}
\subsection{Single-Depth Datasets}
\begin{figure*}[!h]
\begin{center}
\centerline{\includegraphics[width=\textwidth]{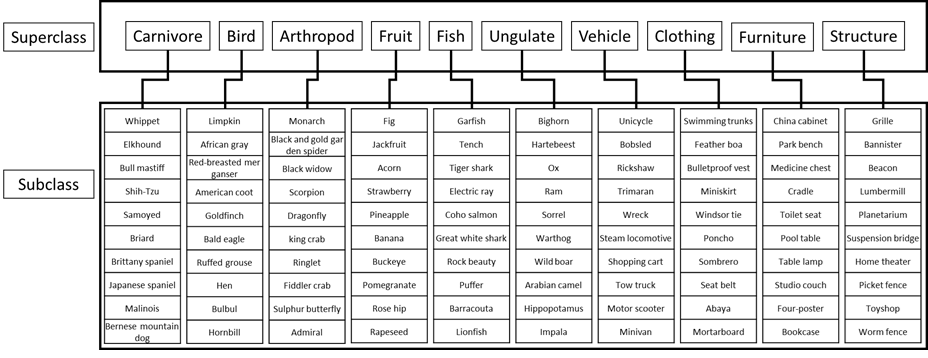}}
\caption{
An overview of the ImageNet-Hier100 dataset, which is represented as a hierarchical structure. The dataset includes 10 broad categories or "superclasses", which are further divided into 10 more specific categories or "subclasses". The subcategories are visually depicted as branches stemming from the main categories, resulting in a total of 100 subclasses in the dataset. 
}
\label{figure6:imagenet depth1 data}
\end{center}
\vskip -0.4in
\end{figure*}

\textbf{ImageNet-Hier100}
ImageNet-Hier100 is a subset of the ImageNet dataset~\cite{deng2009imagenet} that is organized based on the taxonomy of WordNet~\cite{miller1998wordnet}. The dataset is structured to represent 100 fine-grained classes by grouping them into 10 coarse-grained classes that capture the overall semantic structure. Each coarse-grained class consists of 10 corresponding fine-grained classes, which are subcategories that belong to the larger, upper-level group of the coarse-grained class. These 10 coarse-grained classes are referred to as superclasses and include carnivore, bird, arthropod, fruit, fish, ungulate, vehicle, clothing, furniture, and structure. The fine-grained classes, referred as subclasses, that correspond to each superclass are depicted in Figure~\ref{figure6:imagenet depth1 data}.

\subsection{Multiple-Depth Datasets}
\textbf{CIFAR100}
The CIFAR100 dataset is a widely used benchmark for image classification tasks, consisting of 60,000 32x32 color images organized into 100 fine-grained classes, with 600 images per class. The dataset is split into 50,000 training images and 10,000 testing images, making it ideal for evaluating different machine learning models. The images depict real-world objects such as animals, vehicles, and household items and are used for various computer vision tasks, including object recognition and image classification.

For the CIFAR100 dataset, we follow the hierarchical taxonomy as described in ~\cite{garnot2020leveraging}. The dataset has five levels of hierarchy with (2,4,8,20,100) classes, excluding the root node. The dataset has an Imbalance Ratio (IR) of 1, indicating a balanced distribution of classes. It has a total of 134 nodes and 100 leaves, denoting the total number of nodes and leaf nodes in the tree-shaped hierarchy, respectively. The Average Branching Factor (ABF) of the dataset is 3.8, representing the average number of children (subclasses) for each superclass. The average pairwise distance is 7.0, reflecting the average distance between each pair of classes in the hierarchy. 

\textbf{iNaturalist-19}
The iNaturalist dataset is a large-scale image classification dataset of organisms, containing over 800,000 images from more than 8,000 different species. The iNaturalist-19 dataset, a subset of the larger iNaturalist dataset, was introduced for the 2019 CVPR Fine-Grained Visual Categorization Workshop. It includes metadata with hierarchical relationships between species, making it useful for evaluating methods for fine-grained visual categorization. The iNaturalist-19 dataset comprises 265,213 color images, organized into 1010 fine-grained classes.

However, the test set labels for the iNaturalist-19 dataset are not publicly available. To address this issue, we randomly selected and resampled three splits from the original training and validation data to create a new training, validation, and test set, as suggested in~\cite{bertinetto2020making}. These sets were created using probabilities of 0.7, 0.15, and 0.15, respectively.

The iNaturalist-19 dataset has a hierarchical taxonomy with seven levels and (3, 4, 9, 34, 57, 72, 1010) classes, excluding the root node. It also has an Imbalance Ratio (IR) of 31, indicating a significant imbalance in the distribution of classes. The dataset has a total of 1,189 nodes and 1,010 leaves, denoting the total number of nodes and leaf nodes in the hierarchy, respectively. Its Average Branching Factor (ABF) is 6.6, representing the average number of children (subclasses) for each superclass. The average pairwise distance is 11, reflecting the average distance between each pair of classes in the hierarchy.

\section{Details on Implementation of Baseline Methods on HLE setup}
\noindent\textbf{ER and EWC++.} ER and EWC++ uses reservoir sampling strategy for memory management by randomly removing samples in the memory to replace samples. We implemented the reservoir sampling in the hierarchical label expansion (HLE) setup so that it not only ignores the class information but also the hierarchical information when it randomly selects samples to remove from the memory.

\noindent\textbf{BiC.} BiC was originally proposed on the offline CL setup with herding selection~\cite{rebuffi2017icarl}. However, herding selection is not applicable since entire task data is required for computing class mean, which is impossible on online CL setup. Therefore, we applied the reservoir sampling for BiC as used in ~\cite{koh2022online}. BiC empirically demonstrates that the classifier is biased towards new classes and proposes a bias correction layer attached at the end of the classifier, which is trained with the separate validation set as a small part of the memory, to correct the classifier. Since there are multiple classifiers for each hierarchy in the HLE setup, we also used multiple bias correction layers for each corresponding classifier. In contrast, the validation set stores the samples for all encountered classes regardless of their hierarchy.

\noindent\textbf{MIR.} 
MIR enhances memory utilization by first drawing a subset of the memory whose cardinality is larger than that of the training batch, and then selecting samples from the subset that would experience the highest loss increase if trained with streamed data to update the model. To apply MIR in the HLE setup, we extracted samples independently for each hierarchy level and ensured that the ratio of the number of samples in the subset and the training batch matched for each level.

\noindent\textbf{RM, GDumb, and CLIB} RM, GDumb, and CLIB are originally managed to maintain the balance of the number of samples for each class in memory. Following their memory management schemes, we balance it regardless of hierarchy level in the HLE setup.


\begin{figure}[t!]
\begin{center}
\centerline{\includegraphics[width=\columnwidth]{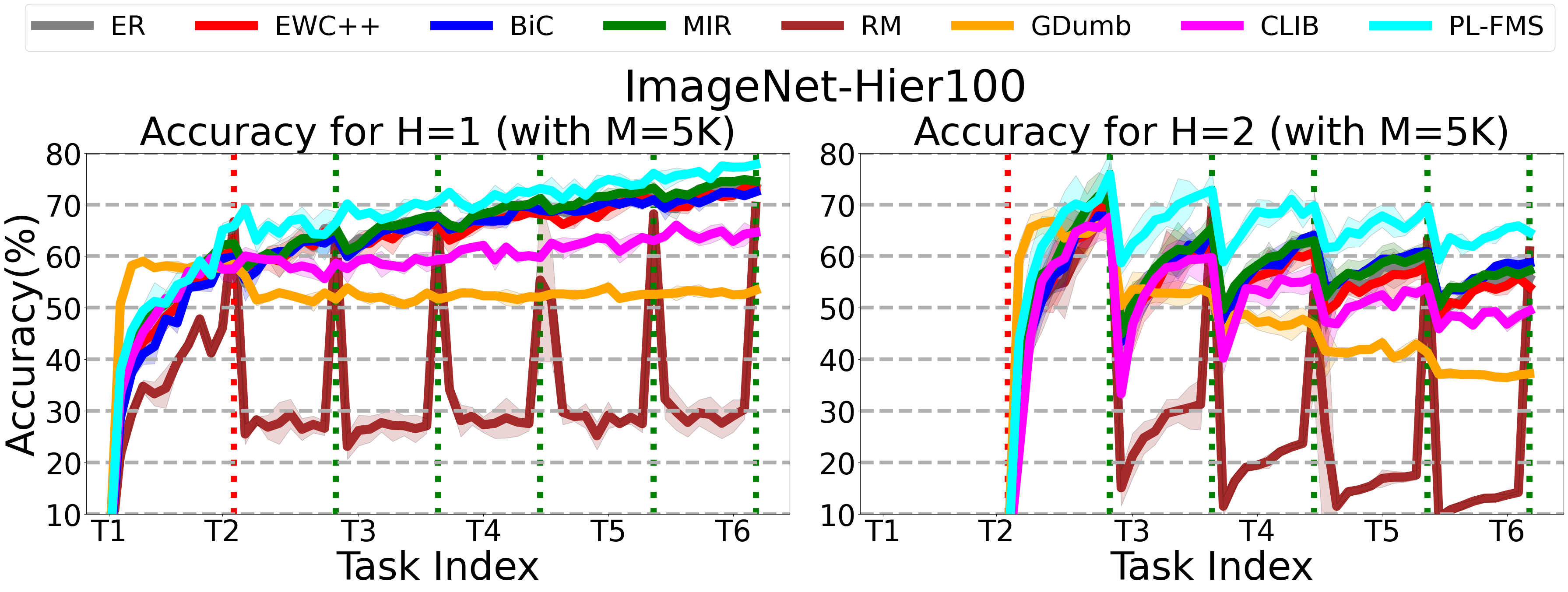}}
\caption{
Any-time inference results on ImageNet-Hier100 for single-label scenario with single-depth hierarchy. H=1 is parent classes and H=2 child classes. Task index 1 receives parent class labeled data and subsequent indexes receive child class labeled data. Each data point shows average accuracy over three runs ($\pm$ std. deviation).
}
\label{figure7:imagenet_hier100_single_label}
\end{center}
\vskip -0.4in
\end{figure}

\section{Details on Evaluation Metrics}
\textbf{Any-time inference}
While average accuracy ($A_{\text {avg}}$) is a widely used measure in continual learning evaluation, it only provides a limited evaluation of a model's performance. $A_{\text {avg}}$ measures performance only at task transitions, which typically occur only a few times during the learning process. Therefore, it may not provide a comprehensive evaluation of a model's ability to adapt to new tasks without forgetting previously learned ones.

In contrast to average accuracy, any-time inference is a more appropriate and useful metric for evaluating continual learning models. Any-time inference measures a model's ability to make accurate predictions at any point during the learning process, without relying on explicit task boundaries. To measure any-time inference, we evaluate the model's accuracy after observing every $\Delta n$ samples, instead of only at discrete task transitions by referring to ~\cite{koh2022online}. This approach provides a more continuous and fine-grained evaluation of a model's performance, reflecting real-world scenarios where new tasks and data can arrive at any time, and the model needs to adapt quickly without sacrificing performance on previously learned tasks. Therefore, any-time inference is a more suitable metric for evaluating continual learning models, as it aligns with the practical requirements of real-world applications where machine learning models must continuously learn and adapt to new data over time.

\begin{figure}[t!]
\begin{center}
\centerline{\includegraphics[width=\columnwidth]{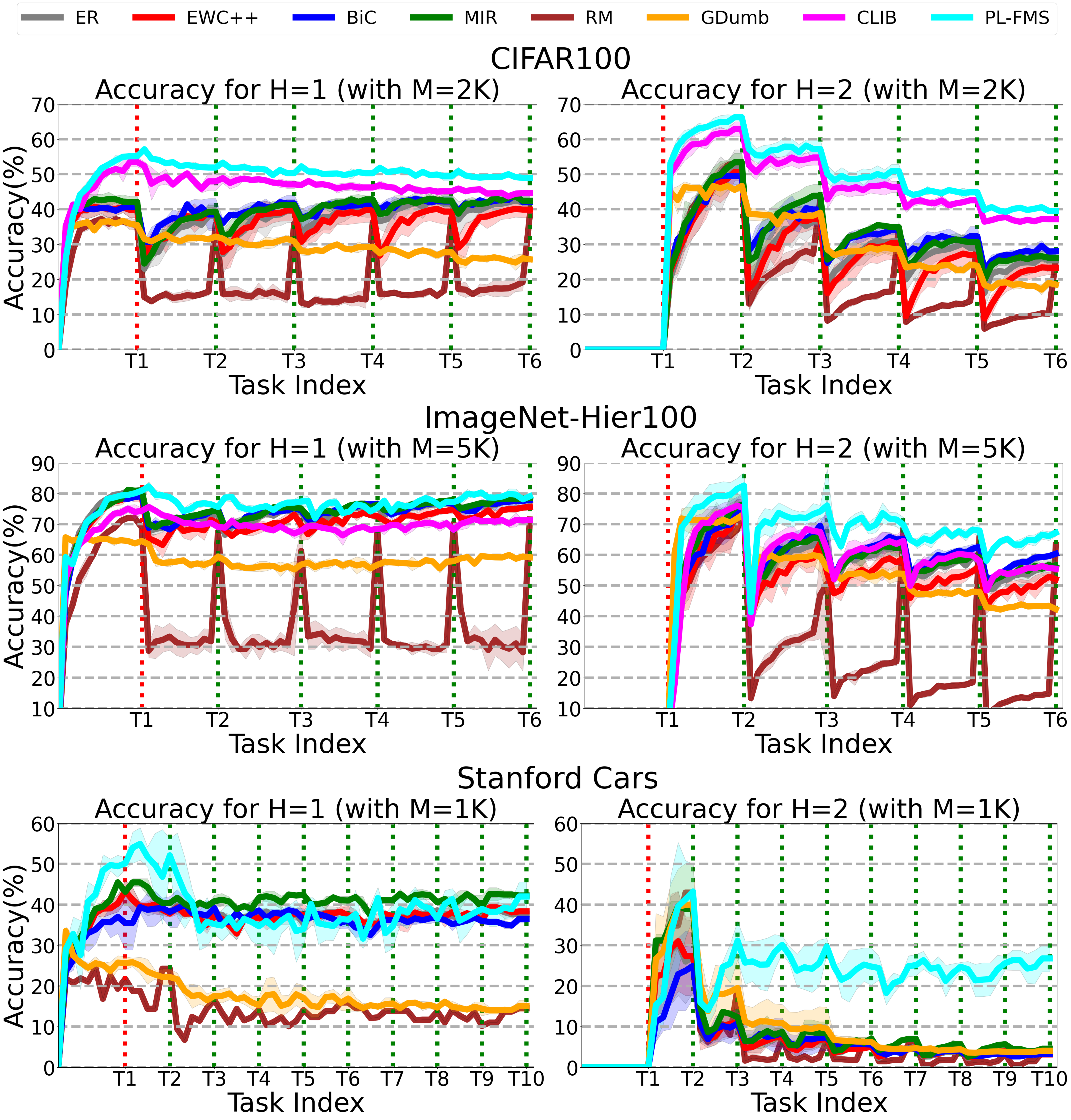}}
\caption{
Any-time inference results on CIFAR100, ImageNet-Hier100, and Stanford Cars dataset for dual-label scenario with single-depth hierarchy. H=1 is parent classes and H=2 child classes. Task index 1 receives parent class labeled data and subsequent indexes receive child class labeled data. Each data point shows average accuracy over three runs ($\pm$ std. deviation).
}
\label{figure8:dual_label}
\end{center}
\vskip -0.4in
\end{figure}

\begin{figure*}[ht!]
\begin{center}
\centerline{\includegraphics[width=\textwidth]{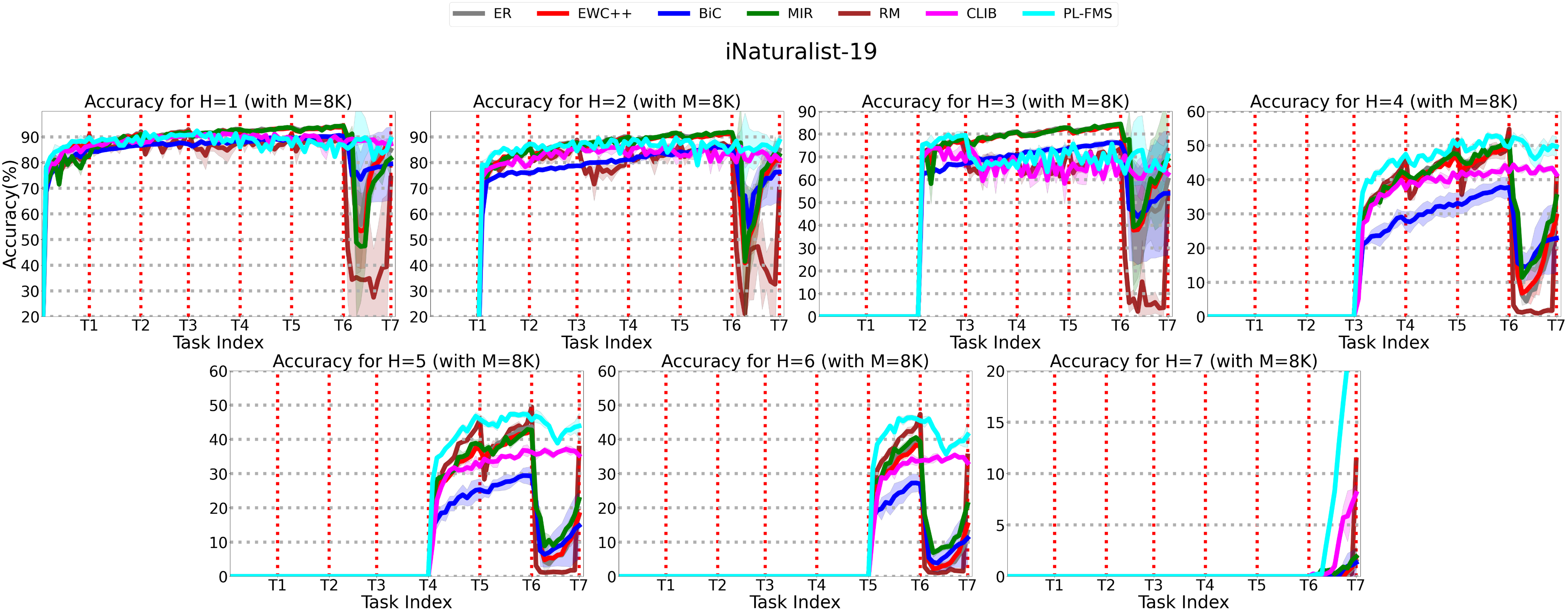}}
\caption{
Any-time inference results on iNaturalist-19 dataset for multiple-depth hierarchy. H=1 represents the coarsest level and H=7 represents the finest level of class hierarchy. The dotted line represents the point at which the model is fully given the task data for the corresponding task index. The reported data points represent the average accuracy over three runs ($\pm$ std. deviation)
}
\label{figure9:inaturalist19}
\end{center}
\vskip -0.4in
\end{figure*}

\begin{figure*}[ht!]
\begin{center}
\centerline{\includegraphics[width=\textwidth]{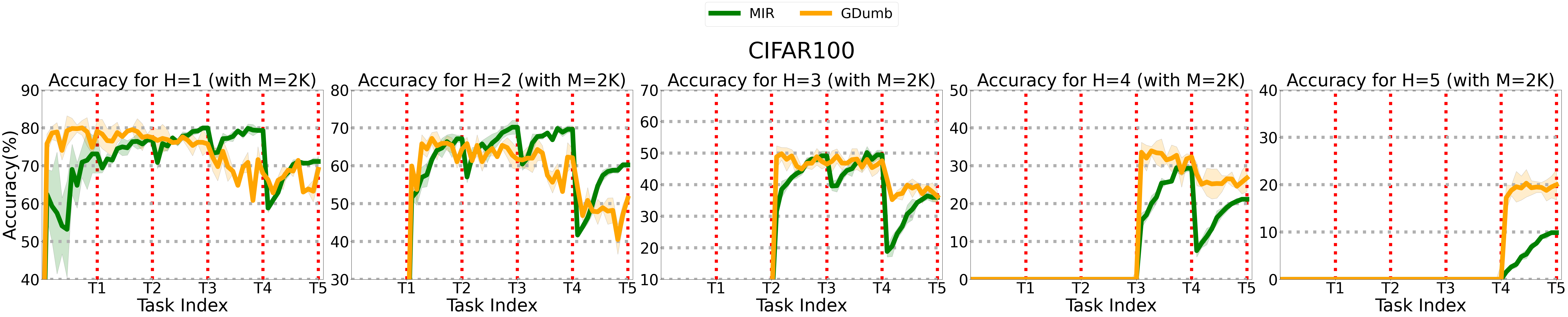}}
\caption{
Any-time inference results on CIFAR100 dataset for multiple-depth hierarchy. H=1 represents the coarsest level and H=5 represents the finest level of class hierarchy. The dotted line represents the point at which the model is fully given the task data for the corresponding task index. The reported data points represent the average accuracy over three runs ($\pm$ std. deviation)
}
\label{figure10:cifar100_mir_gdumb}
\end{center}
\vskip -0.4in
\end{figure*}

\section{Anytime Inference on ImageNet-Hier100 for Single-Label Scenario}
In Figure ~\ref{figure7:imagenet_hier100_single_label}, we report the any-time inference result for ImageNet-Hier100 dataset for single-label scenario with single-depth hierarchy. The trend is similar to the result on CIFAR100 dataset for the single-label scenario in the main paper. It's worth noting that the performance of CLIB was relatively inferior to the methods that employ reservoir sampling, such as ER, EWC++, BiC, and MIR. This could be due to the fact that ImageNet-Hier100 has a longer interval of iterations for each task compared to CIFAR100, and CLIB's memory-only training is limited in its ability to adapt to newly encountered classes.

\section{Anytime Inference on iNaturalist-19 for Multi-Depth Scenario}
Figure~\ref{figure9:inaturalist19} shows the any-time inference results for multi-depth scenario on iNaturalist-19 dataset. As we observed from the any-time inference results for CIFAR100, the performance of the baseline methods except CLIB for the hierarchy levels from 1 to 6 deteriorate seriously at the end of task 6, where the number of class increases explosively from 179 to 1189 by label expansion to the most fine-grained classes. On the other hand, the performance of PL-FMS and CLIB demonstrates their mild forgetting at the end of the task 6 while the any-time inference results of CLIB for the hierarchy levels larger than 4 shows relatively lower performance compared to PL-FMS and RM. Until the end of task 6, EWC++ shows the highest performance for hierarchy level 1,2, and 3, but it exhibit severe catastrophic forgetting after the task 6. In overall, PL-FMS shows the best performance for all hierarchy levels except the hierarchy level 1 at the end of the training and consistently high performance in terms of any-time inference for all hierarchy levels. 
We chose not to conduct the GDumb method for the multiple-depth scenario due to its consistently low performance on both single-depth and multiple-depth datasets, and because it required a significant amount of training time. However, we did perform an additional experiment for the MIR baseline method to clarify the performance of all baseline methods except for GDumb. The MIR method demonstrated comparable performance against other baseline methods, which was the motivation for conducting this experiment.

\section{Anytime Inference for Dual-Label Scenario}

In Figure ~\ref{figure8:dual_label}, we report the any-time inference results for dual-label scenario with single-depth hierarchy on CIFAR100, ImageNet-Hier100, and Stanford Cars dataset. Since the model is trained with more samples and the labels from both hierarchy level 1 and 2 are assigned to same samples, the dual-label scenario showed higher performance compared to the results for the single-label scenario, except some baseline methods. Note that the performance for hierarchy level 1 in dual-label scenario can be more easily saturated in the first task due to the larger number of samples for each task. Because of this, PL-FMS showed the forgetting on CIFAR100 dataset in hierarchy level 1 through the subsequent tasks, while we didn't observe it in the single-label scenario. Furthermore, PL-FMS showed significantly higher performance on Stanford Cars dataset in hierarchy level 2 whereas baseline methods didn't show such dramatic improvement.


\section{Anytime Inference of MIR and GDumb on Mutli-Depth Scenarios}
 Figure \ref{figure10:cifar100_mir_gdumb} is the results of the any-time inference of MIR and GDumb for multi-depth scenario on CIFAR100 dataset. We can find that MIR shows similar performance to ER and EWC++. Also, GDumb exhibits similar trend that we found from the single-depth scenario, where the performance is maintained during the task since it is trained from the scratch whenever the model is tested. Because of that, as can be seen from the result of RM at the end of the last task, GDumb shows relatively higher performance for the highest hierarchy level compared to other baselines. This is due to the fact that both RM and GDumb train the model with samples in memory for multiple epochs, which is not realistic for task-free online continual learning.

\end{document}